**Title:** Biologically-inspired neuronal adaptation improves learning in neural networks.

**Authors:** Yoshimasa Kubo*, Eric Chalmers, Artur Luczak*

**Address:** Canadian Centre for Behavioural Neuroscience, University of Lethbridge, AB, Canada

*****Correspondence:** yoshi.kubo@uleth.ca, luczak@uleth.ca

**Abstract**:

Since humans still outperform artificial neural networks on many tasks, drawing inspiration from the brain may help to improve current machine learning algorithms. Contrastive Hebbian Learning (CHL) and Equilibrium Propagation (EP) are biologically plausible algorithms that update weights using only local information (without explicitly calculating gradients) and still achieve performance comparable to conventional backpropagation. In this study, we augmented CHL and EP with *Adjusted Adaptation,* inspired by the adaptation effect observed in neurons, in which a neuron's response to a given stimulus is adjusted after a short time. We add this adaptation feature to multilayer perceptrons and convolutional neural networks trained on MNIST and CIFAR-10. Surprisingly, adaptation improved the performance of these networks. We discuss the biological inspiration for this idea and investigate why Neuronal Adaptation could be an important brain mechanism to improve the stability and accuracy of learning.

**Introduction**:

Deep neural networks outperform humans in Atari games [1] and the Game of GO [2] , but fall short of humans in tasks such as art, music, and translations. Thus, looking for inspiration from the brain may help to improve current deep neural networks. Many researchers are drawing inspiration from the brain to close the gap between biological and machine learning. For example, while the backpropagation algorithm (BP) has long been used to train deep neural networks by backpropagating error signals through the layers of the network [3], the debate over whether biological neurons could support such an operation [4, 5], has led to several more biologically plausible algorithms being proposed [6-12]. This paper will focus on two of these: Contrastive Hebbian learning (CHL) [13-15] and the closely related Equilibrium propagation (EP)[12, 16-18]. These algorithms model the network as a dynamical system, and learn from temporal differences in activity rather than explicitly calculated errors or gradients.

CHL and EP consists of two learning phases: a free or negative phase, followed by a clamped or positive phase. During the free phase, an input signal is presented and the network is allowed to equilibrate to a steady-state. The clamped phase then clamps output neurons to the desired targets, and the network is allowed to re-equilibrate. CHL clamps output neurons completely, while EP uses "soft" clamps, and nudges the output activities toward the desired level - which may be more biologically plausible. Weight updates are based on the differences between free- and clamped-phase activities of neurons on either side of the weight. Our previous study [19] showed that free-phase steady-state activity can be well predicted based on the first few steps of neural dynamics. This would mean that a full free-phase equilibration may not actually be required before output clamping, and that learning could occur without two distinct phases - making the EP concept even more biologically plausible.

In this study, we add *adjusted adaptation* to CHL and EP. This is based on the *Neural Adaptation* effect that has been observed in biological neurons in several systems [20, 21]. During neural adaptation, a neuron's response to a given stimulus will usually decrease over a short period of time from its initial level. This could be interpreted as a smooth change of sensitivity to the stimulus. Here we model adaptation during the clamped phase, such that each neuron's clamped-phase activity is gradually pushed back toward its free-phase activity. This reduces the difference between the free- and clamped-phase activations; modulating weight updates in a way that makes training smoother.

This paper extends our previous work [19, 22] by applying the adaptation concept to multilayer perceptrons and convolutional neural networks trained on MNIST [23] and CIFAR-10 [24] by CHL and EP. We make the following contributions:

1. Demonstrate that models with adaptation achieve better performance than models without adaptation on both MNIST and CIFAR-10 tasks.
2. We provide an explanation of why adaptation may work by comparing the training gradients created with and without adaptation and comparing these to the gradients created by backpropagation.

**Method**:

In this section, we discuss some notations, methods, and model specifications for our experiments.

*Equilibrium Propagation (EP) and Contrastive Hebbian Learning (CHL):*

During the free phase of EP, the network calculates the dynamics of activity without any target signals or gradient signals. In a one-hidden-layer multilayer perceptron, the equations of dynamics for activity at each layer are described based on previous work [12, 16] as:

$$x_{o,t} = x_{o,t-1} + h\left(-x_{o,t-1} + p(\Sigma_j \, w_{j,o} \, x_{j,t-1} + b_o)\right), \tag{1}$$

$$x_{j,t} = x_{j,t-1} + h\left(-x_{j,t-1} + p(\Sigma_i \, w_{i,j} \, x_{i,t-1} + \gamma \Sigma_o \, w_{o,j} \, x_{o,t-1} + b_j)\right), \tag{2}$$

where $x$ is an activity, $w$ represents weights for each layer, $i, j$, and $o$, are indexes of input, hidden and output layer neurons, and $b$ is a bias. $p$ is an activation function such as the sigmoid function, and $\gamma$ is the feedback parameter. $h$ is the Euler method's time-step.

During the clamped phase, the output neurons are influenced by target signals. During this phase the network calculates the dynamics of activity in the output layer as:

$$x_{o,t} = x_{o,t-1} + h\left(-x_{o,t-1} + p(\Sigma_j \, w_{j,o} \, x_{j,t-1} + b_o) + \beta(y - x_{o,t-1})\right), \tag{3}$$

where y is a target signal, and $\beta$ is a nudging parameter that pushes output-layer activations back toward their free-phase level.

Given these activations at the free and clamped phases, weights will be updated by

$$\Delta w_{pre,post} = \frac{1}{\beta}\alpha\left(\hat{x}_{pre}\hat{x}_{post} - \check{x}_{pre}\check{x}_{post}\right), \tag{4}$$

where $\hat{x}$ is an activity at the clamped phase, $\check{x}$ is an activity at the free phase, $\alpha$ is the learning rate. *pre* and *post* are previous and post layer neuron indexes, respectively (e.g. for $\Delta w_{\{i,j\}}$, *pre* and *post* will be i and j, respectively).

Contrastive Hebbian learning (CHL) is very similar to EP. Instead of Equations 3 and 4, CHL uses:

$$x_{o,t} = y, \tag{5}$$

$$\Delta w_{pre,post} = \alpha \left( \hat{x}_{pre}\hat{x}_{post} - \check{x}_{pre}\check{x}_{post} \right) \tag{6}$$

Note that Equation 5 is used only for the clamped phase at the output layer. The free phase uses Equation 1 at the output layer. Thus, CHL clamps output neurons completely, where EP uses a "soft" clamping effect. The soft clamping weekly nudges the outputs at the free phase to their targets to minimize the difference between current outputs and targets. On the other hand, CHL uses a "hard" clamping, where the output neurons are clamped at the desired target value.

Note that when gaps between $\hat{x}$ and $\check{x}$ are large, especially, at the top layer, $\Delta w$ would also be large - potentially leading to abrupt jumps in network weights. Softening the clamping by reducing $\beta$ does not solve this problem, because the $1/\beta$ term in Equation 4 amplifies the differences. We can reduce $\alpha$, but this slows learning convergence [25, 26], and a smaller learning rate could lead the network to find sharp local minima [27].

*Predictive Learning Rule:*

In this subsection, we discuss our learning rule. Our rule modifies Equation 6 by replacing $\check{x}_{pre}$ (the free-phase activity of the presynaptic neuron), with $\hat{x}_{pre}$ (the *clamped-phase* activity of the presynaptic neuron). Our previous work [19] showed that a rule of this form emerges naturally if we assume that each neuron is working to maximize its metabolic energy. The new rule is as follows:

$$\Delta w_{pre,post} \propto \frac{1}{\beta} \alpha (\hat{x}_{pre}\hat{x}_{post} - \hat{x}_{pre}\check{x}_{post}) = \frac{1}{\beta} \alpha\, \hat{x}_{pre}(\hat{x}_{post} - \check{x}_{post}), \tag{7}$$

In case of CHL, $\frac{1}{\beta}$ is removed. We call this update rule the *predictive learning rule* because our previous study used *predicted* free-phase steady-state activity ($\tilde{x}$) in place of $\check{x}$ (allowing neurons to predict their own free-phase steady state may be more biologically plausible, as it allows learning to occur without requiring two distinct phases). However, for the purpose of investigating adaptation, this study computes free-phase steady-state activities in the conventional way - which is the same as assuming perfect predictions of free-phase activity (a reasonable assumption, as our previous work found correlation between predicted ($\tilde{x}$) and actual free phase activity ($\check{x}$) was R=1 ± 0.0001 SD [19]). For consistency with this previous study [19], we apply delay to a clamped (teaching) signal to models trained by CHL. For example, Figure 1 shows that the signal is clamped after 12 steps. In our previous paper, we discuss that such delay could be more biologically-plausible, as in the visual cortex a top-down "teaching" signal (similar to clamped phase) arrives tens of ms later than the initial bottom-up signal (similar to the free phase).

*Adjusted Adaptation:*

*Adjusted Adaptation* was introduced by our previous study [22], based on neural adaptation observed in the brain. To implement *Adjusted Adaptation*, the activities at the clamped phase are nudged toward activities at the free phase to reduce the gap between these activities. We speculate that these smaller gaps give smoother weight updates compared to conventional EP. We model adaptation as follows:

$$\hat{x}_{adp,t+1} = (1-c)\hat{x}_{adp,t} + c * \check{x}_*, \tag{8}$$

where $c$ is a coefficient parameter, $\check{x}_*$ is a steady-state at the free phase. Equation 8 is applied during additional steps after computing the clamped-phase dynamics. We describe this algorithm as pseudocode next to Figure 1. Figure 1 depicts the adapted dynamics of activities during a CHL clamped phase. In this case, the clamped (teaching) signal is given 12 steps after the input signal is presented, as explained earlier. Adaptation is applied after 120 steps, and nudges the clamped phase activity back toward free-phase activity.

*Model specifications:*

For multilayer perceptron (MLP) with CHL on the MNIST dataset, the number of time steps for the free phase and clamped phases was set to 120. We tried seven different architectures: 782-6-10 (meaning the number of neurons at input – hidden – output layers), 782-50-10, 782-50-10 with lateral connection on the hidden layer, 782-50-10 with lateral connection on both the hidden and top layer, 782-1000-10, 782-1000-10 with lateral connection on the hidden layer, 782-1000-10 with lateral connection on both the hidden and top layer. We conducted these experiments with different architecture to check that we still get robust results on our models even if we change parameters. We used the learning rate of 0.1 on 782-6-10 and 782-50-10 models, and learning rate 0.03 for 782-1000-10 models. The teaching signal delay was 12 steps and $h$ was 0.1 for all the networks. For our model with the adjusted adaptation, extra steps for adjusted adaptation are 20 and $c$ was 0.1. All models use the sigmoid activation function. For all experiments we used the AdaGrad optimizer [28] (to find how to implement Contrastive Hebbian Learning with AdaGrad please see our code at https://github.com/ykubo82/bioCHL/blob/add-license-1/CHL_clamped.py, specifically Line 107).

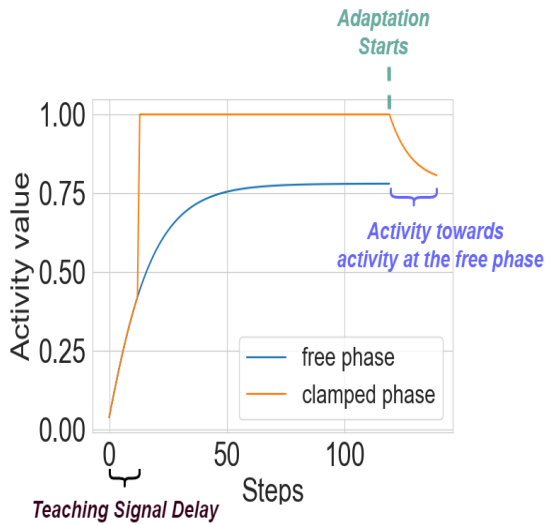

**Figure 1:** An example of neuron activity during the free and clamped phases with adjusted adaptation. For visual clarity, only one representative neuron from the top layer is shown. The time steps for free and clamped phases are 120, and teaching signals are given after 12 steps. The adjusted adaptation steps are 20. After 120 steps at the clamped phase, additional 20 steps for the adaptation are applied using Eq. 8.

**Algorithm 1** Basic algorithm to train our models.

**Input:** Data $(X, Y)$, synaptic weights $w$, biases $b$, Step Size $T$, adaptation step size $A$, Learning Epoch $E$, Learning Rate $\alpha$, Index of Pre-Synaptic Neuron $i$, Index of Post-Synaptic Neuron $j$, Index of Output layer neurons $o$, Time step $dt$, Activation Function $p$

**for** $i = 1, 2, \ldots, E$ **do**
    $x_0, y_0 \leftarrow SampleMiniBatchData(X, Y)$
    **for** $t = 1, 2, \ldots, T$ **do**
        $\check{x}_{i,t-1} \leftarrow x_0$ ▷ Clamp inputs
        Compute $\check{x}$ with Eqs 1 and 2
    **end for**
    **for** $t = 1, 2, \ldots, T + A$ **do**
        $\hat{x}_{i,t-1} \leftarrow x_0$ ▷ Clamp inputs
        $\hat{x}_{o,t-1} \leftarrow y_0$ ▷ Clamp teaching signals
        Compute $\hat{x}$ with Eqs 1 and 3 or 5
        **if** $t > T$ **then**
            Compute $\hat{x}$ with Eqs 8
        **end if**
    **end for**
    //Modify synaptic weights
    Compute $\Delta w_{i,j}$ with Eq 7
    Compute $\Delta w_{j,o}$ with Eq 7
    $w_{i,j} \leftarrow w_{i,j} + \Delta w_{i,j}$
    $w_{j,o} \leftarrow w_{j,o} + \Delta w_{j,o}$
**end for**

For the convolutional neural networks (CNN) with EP on the CIFAR-10 dataset, the time steps for the free phase and clamped phases are 130 and 30, respectively, and $\beta$ is 0.18. We set the learning rates for the network to 0.21, 0.021, and 0.021 for the first, second convolutional layer, and fully connected layer, respectively. Our model consists of 256 and 512 filters whose sizes are 3x3 for both 1st and 2nd layers and followed by one dense layer for the output layer. $h$ is 1.0 for this model. For the clamped phase, activities from the free phase at time step 110 are used as the initial activities (this is the same as teaching signal delay). This can be seen as a delay similar to our previous experiments. For our model with the adjusted adaptation, extra steps for implementing adaptation are 10 and $c$ is 0.1. The activation function for these models is the hard sigmoid function [17]. We did not use any optimizer for these CNN models. Our CNN models are based on [16] and [22].

Code for our networks showing all the implementation details is available at: https://github.com/ykubo82/AdpNet

**Results:**

Table 1 and Figure 2 shows the results for MLP with/without the adaptation on the MNIST dataset. We found that adaptation improved performance, for all tested models. Similarly, convolutional neural networks with the adjusted adaptation (CNN-ADP) on CIFAR-10 achieved a better test error of 19.51 ± 0.6% as compared to the model without the adjusted adaptation (CNN, a test error of 22.46 ± 0.59%). Figure 3 shows the learning curves for these models. Note that the learning curves for models with adaptation are smoother. We consider this smoothness to be the result of adaptation, which reduces the gap between activities at the free and clamped phase.

**Table 1.** Training results on MNIST and CIFAR-10 with EP and the adaptation/no adaptation. For MNIST, we trained multilayer perceptron with CHL. For CIFAR-10, we trained convolutional neural networks with EP. For *laterl1*, the lateral connections are only in the hidden layer. For *lateral2*, the lateral connections are in both hidden and output layers. We trained each network 6 times to calculate average and ± the standard deviation.

|  | No Adaptation (error %) | | Adaptation (error %) | |
|---|---|---|---|---|
|  | Test | Train | Test | Train |
| MNIST (MLP) | | | | |
| 784-6-10 | 17.29 ± 1.78 | 16.11 ± 1.77 | **11.93 ± 0.84** | 9.45 ± 1.06 |
| 784-50-10 | 4.63 ± 0.19 | 2.71 ± 0.22 | **3.56 ± 0.1** | 1.62 ± 0.09 |
| 784-50-10; lateral 1 | 4.56 ± 0.30 | 2.54 ± 0.24 | **3.82 ± 0.13** | 1.76 ± 0.12 |
| 784-50-10; lateral 2 | 4.88 ± 0.33 | 2.81 ± 0.36 | **3.86 ± 0.18** | 1.84 ± 0.11 |
| 784-1000-10 | 1.98 ± 0.04 | 0.00 ± 0.00 | **1.77 ± 0.06** | 0.01 ± 0.00 |
| 784-1000-10; lateral 1 | 1.85 ± 0.05 | 0.00 ± 0.00 | **1.78 ± 0.06** | 0.01 ± 0.00 |
| 784-1000-10; lateral 2 | 1.83 ± 0.06 | 0.00 ± 0.00 | **1.81 ± 0.07** | 0.01 ± 0.00 |
| CIFAR10 (CNN) | | | | |
| 256-512 | 22.46 ± 0.59 | 22.88 ± 1.71 | **19.51 ± 0.6** | 14.70 ± 2.31 |

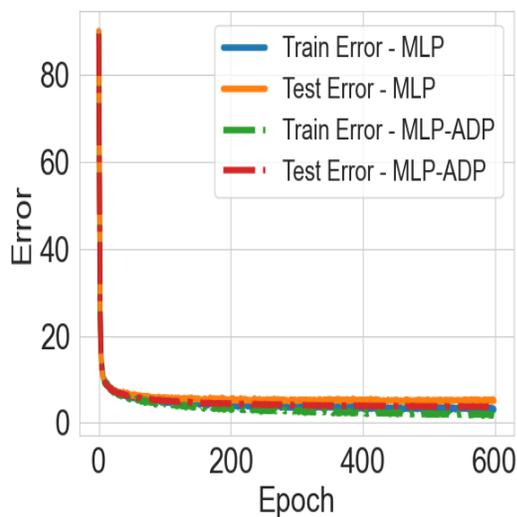
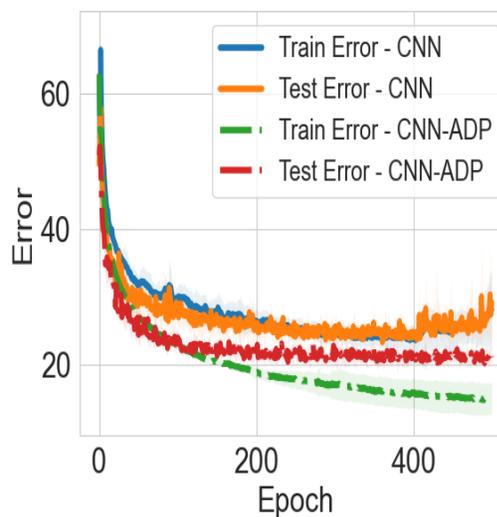

**Figure 2**. The learning curves for the models (parameters:782-50-10) with the adjusted adaptation (MLP-ADP) and without the adjusted adaptation (MLP) on MNIST.

**Figure 3.** The learning curves for the models with the adjusted adaptation (CNN-ADP) and without the adjusted adaptation (CNN) on CIFAR-10.

*Gradient Checks:*

Why does adaptation work? To answer this question, we calculated the angles for the weights' gradients between CHL models with/without the adaptation and models trained with backpropagation on the MNIST dataset. These gradients were calculated without applying the optimizer. This comparison is inspired by Lilicrap et al. [29]. Specifically, we calculated the angle θ between two weight gradient vectors, *a* and *b*:

$$\theta = cos^{-1}\left(\frac{||a \cdot b||}{||a|| \cdot ||b||}\right) \tag{9}$$

where each value in vector *a* represents a single weight update calculated as explained above: $\Delta w_{\{pre,post\}} = \alpha\, \hat{x}_{pre}(\hat{x}_{post} - \check{x}_{post})$. Similarly, corresponding values in vector *b* represent change in weights calculated using BP method, in the same network. For computational efficiency, we used a network with only 6 neurons in a hidden layer. We found that weights' gradients calculated in a network with adaptation were closer to BP gradients, than gradients without adaptation (Figures 4). This suggests that adaptation changes network dynamics in a way that makes weight gradients closer to "optimal".

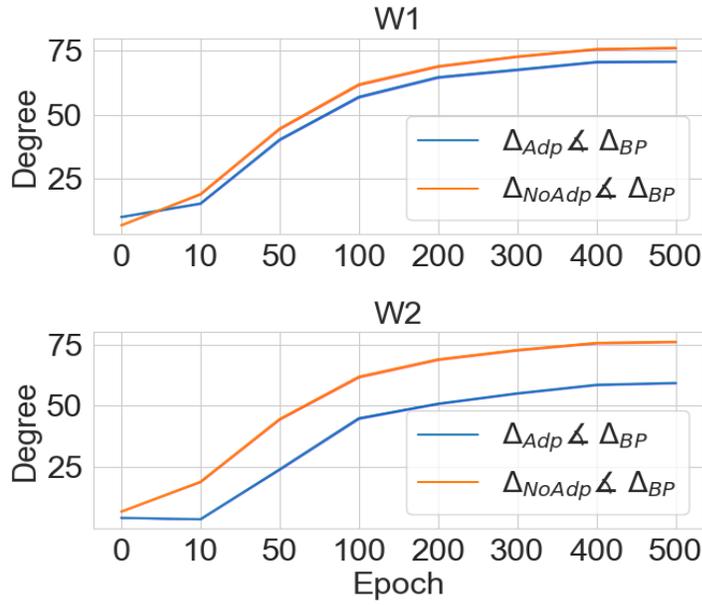

**Figure 4.** Mean angle between weight gradients calculated using BP and CHL. Blue line shows the angle calculated for CHL with adaptation, and orange for CHL without adaptation. "W1" means the gradients for the weights between the input and hidden layers, and "W2" means the gradients for the weights between the hidden and output layers. The angle between the gradients with the adaptation and BP is denoted as $\Delta Adp \measuredangle \Delta BP$. The angle between the gradients without the adaptation and BP is denoted as $\Delta NoAdp \measuredangle \Delta BP$. Note that the blue line is consistently closer to 0, which demonstrates that CHL with adaptation provides weights updates more similar to BP gradients.

For sanity check, we also tested that implementing adaptation is not simply equivalent to reducing a learning rate. To test it we trained MLP (782-6-10) without the adaptation but with learning rates reduced to (1) 0.05, (2) 0.01, and (3) 0.001. We trained these models until the test error started increasing. The smallest test errors for networks with the learning rates (1), (2), and (3) were 16.03±1.38%, 13.95±0.39%, and 13.22±1.25% respectively. This shows that networks without adaptation and with smaller learning rates still have larger test errors as compared to the network with adaptation (11.93 ± 0.84%; learning rate: 0.1).

**Conclusions**:

Neural adaptation has been observed in all types of neurons in vertebrates, as well as in invertebrates [20, 21]. Neuronal adaptation can be defined as a change in activity over time in response to the same sensory stimulus, like sound, light, or tactile stimulation. Usually, neuron activity adapts most rapidly at the beginning, and plateauing at a steady-state value; similarly as implemented in our model (Figure 1). Interestingly, it was also proposed that neuronal adaptation could be a brain mechanism for surprise minimization, which may underlie conscious perception [22]. Considering that neural adaptation is a ubiquitous phenomenon across neuronal systems, thus it may serve an important function in neural information processing. Here we provide the first quantitative account of how neuronal adaptation can improve learning in deep neural networks.

Such improved learning could be due to fact that if activity in the clamped phase is much different from activity without clamp, then learning may deteriorate as those two network states could be in different modes of the energy function [10, 12, 22]. Adaptation may thus reduce this problem by bringing clamped state closer to already learned state (free phase). This could also have a more cognitive interpretation. For that, let us use an analogy: if part of a car is occluded by a tree, then purely on sensory information we cannot say what is behind that tree. However, based on what we learned about the world so far, we know what shape has a car, and thus we can assume that the rest of the car is likely behind the tree. Similarly, neuronal adaptation may allow a network to use already learned information (manifested by activity in free phase) as a strong context to more appropriately store new information (clamped signal).

We also note that adaptation could provide regularization effect: if there is a large difference between a particular neuron's clamped and free-phase activities, that neuron would experience a stronger push back toward free-phase activity. The result would be a reduction of that neuron's effect on the overall network's direction during learning. Thus, adaptation could be seen as a new activation regularization method, similar to the commonly used dropout method [30], which is also an activation regularization.

In future work we plan to implement neuronal adaptation in models trained only with BP. Regularization effect of adaptation may help to improve training networks with BP. This might also help us to better understand the biological function of adaptation.

**Funding**: This study is supported by NSERC DG and CIHR Project grants to AL.

**Conflicts of Interest:** The authors declare that they have no conflicts of interest.

**References:**

1. Mnih, V., et al., *Human-level control through deep reinforcement learning.* nature, 2015. **518**(7540): p. 529-533.

2. Silver, D., et al., *Mastering the game of Go with deep neural networks and tree search.* nature, 2016. **529**(7587): p. 484-489.

3. Rumelhart, D.E., G.E. Hinton, and R.J. Williams, *Learning internal representations by error propagation*. 1985, California Univ San Diego La Jolla Inst for Cognitive Science.

4. Crick, F., *The recent excitement about neural networks.* Nature, 1989. **337**(6203): p. 129-132.

5. Lillicrap, T.P., et al., *Backpropagation and the brain.* Nature Reviews Neuroscience, 2020. **21**(6): p. 335-346.

6.   Bengio, Y., *How auto-encoders could provide credit assignment in deep networks via target propagation.* arXiv preprint arXiv:1407.7906, 2014.

7.   Hinton, G.E. and J. McClelland. *Learning representations by recirculation*. in *Neural information processing systems*. 1987.

8.   Lecun, Y., *PhD thesis: Modeles connexionnistes de l'apprentissage (connectionist learning models).* 1987.

9.   Lillicrap, T.P., et al., *Random synaptic feedback weights support error backpropagation for deep learning.* Nature communications, 2016. **7**(1): p. 1-10.

10.  Movellan, J.R., *Contrastive Hebbian learning in the continuous Hopfield model*, in *Connectionist models*. 1991, Elsevier. p. 10-17.

11.  O'Reilly, R.C., *Biologically plausible error-driven learning using local activation differences: The generalized recirculation algorithm.* Neural computation, 1996. **8**(5): p. 895-938.

12.  Scellier, B. and Y. Bengio, *Equilibrium propagation: Bridging the gap between energy-based models and backpropagation.* Frontiers in computational neuroscience, 2017. **11**: p. 24.

13.  Almeida, L.B., *A Learning Rule for Asynchronous Perceptrons with Feedback in a Combinatorial Environment*, in *Proceedings of the IEEE First International Conference on Neural Networks San Diego, CA*, M. Caudil and C. Butler, Editors. 1987. p. 609-618.

14.  Baldi, P. and F. Pineda, *Contrastive learning and neural oscillations.* Neural computation, 1991. **3**(4): p. 526-545.

15. Pineda, F.J., *Generalization of back-propagation to recurrent neural networks.* Physical review letters, 1987. **59**(19): p. 2229.

16. Ernoult, M., et al., *Updates of equilibrium prop match gradients of backprop through time in an RNN with static input.* Advances in neural information processing systems, 2019. **32**.

17. Laborieux, A., et al., *Scaling equilibrium propagation to deep convnets by drastically reducing its gradient estimator bias.* Frontiers in neuroscience, 2021. **15**: p. 129.

18. Scellier, B. and Y. Bengio, *Equivalence of equilibrium propagation and recurrent backpropagation.* Neural computation, 2019. **31**(2): p. 312-329.

19. Luczak, A., B.L. McNaughton, and Y. Kubo, *Neurons learn by predicting future activity.* Nature Machine Intelligence, 2022: p. 1-11.

20. Benda, J., *Neural adaptation.* Current Biology, 2021. **31**(3): p. R110-R116.

21. Whitmire, C.J. and G.B. Stanley, *Rapid sensory adaptation redux: a circuit perspective.* Neuron, 2016. **92**(2): p. 298-315.

22. Luczak, A. and Y. Kubo, *Predictive neuronal adaptation as a basis for consciousness.* Frontiers in Systems Neuroscience, 2021. **15**.

23. LeCun, Y., et al., *Gradient-based learning applied to document recognition.* Proceedings of the IEEE, 1998. **86**(11): p. 2278-2324.

24. Krizhevsky, A. and G. Hinton, *Learning multiple layers of features from tiny images.* 2009.

25. Li, F.-F et al., *CS231n: Convolutional Neural Networks for Visual Recognition.* 2021 [cited 2021 October 1st]; Available from: https://cs231n.github.io/neural-networks-3/#annealing-the-learning-rate.

26. Sun, J., et al., *A deep learning perspective of the forward and inverse problems in exploration geophysics.* CSEG Geoconvention, 2019.

27. Li, Y., C. Wei, and T. Ma, *Towards explaining the regularization effect of initial large learning rate in training neural networks.* Advances in Neural Information Processing Systems, 2019. **32**.

28. Duchi, J., E. Hazan, and Y. Singer, *Adaptive subgradient methods for online learning and stochastic optimization.* Journal of machine learning research, 2011. **12**(7).

29. Lillicrap, T.P., et al., *Random feedback weights support learning in deep neural networks.* arXiv preprint arXiv:1411.0247, 2014.

30. Srivastava, N., et al., *Dropout: a simple way to prevent neural networks from overfitting.* The journal of machine learning research, 2014. **15**(1): p. 1929-1958.